\documentclass{article}

\usepackage[accepted]{icml2026} 
\usepackage[utf8]{inputenc}
\usepackage[T1]{fontenc}
\usepackage{hyperref}
\usepackage{url}
\usepackage{booktabs}
\usepackage{amsfonts}
\usepackage{amsmath}
\usepackage{nicefrac}
\usepackage{microtype}
\usepackage{graphicx}
\usepackage{multirow}
\usepackage{xcolor}
\usepackage{enumitem}
\usepackage{subcaption}
\usepackage{tikz}
\usetikzlibrary{shapes,arrows,positioning,fit,backgrounds}
\setlist{nosep,leftmargin=*}

\begin{document}

\twocolumn[
\icmltitle{V2V-Bench: A Comprehensive Benchmark for \\ Video-to-Video Generation Evaluation}

\icmlsetsymbol{equal}{*}

\begin{icmlauthorlist}
\icmlauthor{Tao Liu}{equal,centific}
\icmlauthor{Leela Krishna}{equal,centific}
\icmlauthor{Gouti Pavan Kumar}{centific}
\icmlauthor{Sreeja K}{centific}
\icmlauthor{Vishav Garg}{centific}
\end{icmlauthorlist}

\icmlaffiliation{centific}{Centific Global Solutions Inc.}
\icmlcorrespondingauthor{Tao Liu}{tao.liu@centific.com}
\icmlcorrespondingauthor{Leela Krishna}{leela.krishna@centific.com}

\icmlkeywords{video-to-video generation, benchmark evaluation, video generation metrics, temporal consistency, frame-level correspondence, structural fidelity, edit faithfulness, human preference alignment, VLM-as-judge, diffusion video models}

\vskip 0.3in
]

 \printAffiliationsAndNotice{}

\begin{abstract}
Video-to-video (V2V) generation is difficult to evaluate because outputs must both follow editing instructions and preserve frame-level correspondence with the source video, which existing T2V and I2V metrics do not capture. We introduce V2V-Bench, a 11-dimension benchmark organized into five categories: temporal alignment, structural fidelity, transformation quality, video quality, and semantic alignment. V2V-Bench pairs diverse source videos with challenging editing tasks and evaluates two commercial models, Grok Imagine and Gemini Veo3.1, and one open-source model, Open Sora 2. Results show complementary model strengths: Grok performs better on editing fidelity, while Veo3.1 achieves stronger visual quality. On six V2V-specific dimensions, V2V-Bench reaches a Spearman correlation of $0.905$ with human judgments.
\end{abstract}

\section{Introduction}
\label{sec:intro}
Video-to-video generation has become an important paradigm for controllable video editing, where a model is required to transform a source video according to a target instruction while preserving its temporal structure, scene dynamics, and spatial relationships~\cite{wang2018video,jeong2025reangle,hu2023videocontrolnet}. Although recent diffusion-based and autoregressive video models have made substantial progress in generating realistic motion and high-fidelity visual content, evaluating video-to-video generation remains challenging~\cite{hacohen2024ltx,hu2023videocontrolnet}. Existing benchmarks largely emphasize perceptual quality, semantic relevance, or general video realism, but these criteria do not fully capture the core requirement of video-to-video transformation: maintaining fine-grained correspondence with the source video while faithfully applying the intended edit~\cite{huang2023vbench,han2025video,zheng2025vbench,liu2023fetv}.

To address this gap, we introduce the V2V-Bench, a comprehensive benchmark for video-to-video generation. V2V-Bench provides a hierarchical and disentangled evaluation framework covering 11 fine-grained dimensions across temporal alignment, structural fidelity, transformation quality, video quality, and semantic alignment. Instead of reporting only aggregate scores, our benchmark offers interpretable diagnostics that reveal how models succeed or fail under different transformation requirements.
\begin{figure}[!t]
  \centering
  \includegraphics[width=0.5\textwidth]{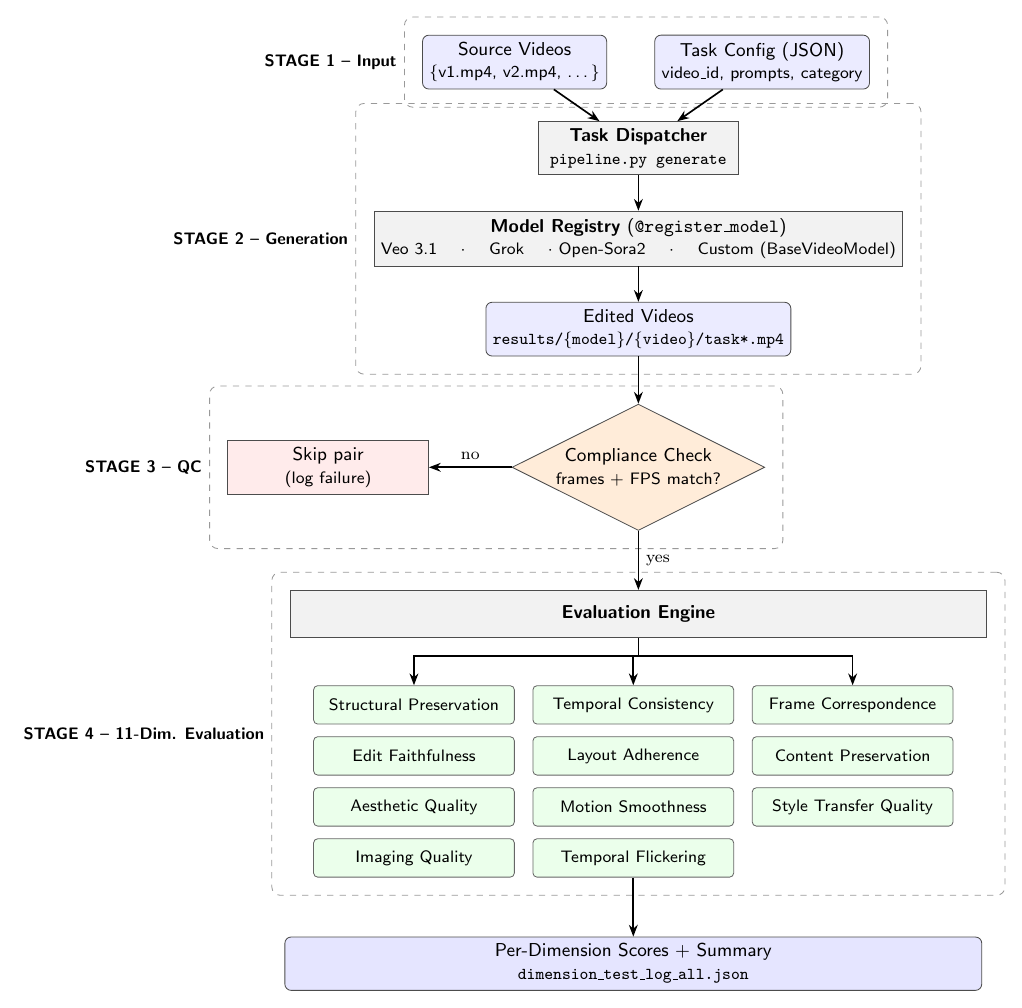}
  \caption{Overview of the V2V-Bench framework.}
  \label{fig:pipeline}
\end{figure}
V2V-Bench includes 81 curated source videos that span diverse scenes, motions, and visual content. Each video is paired with carefully designed editing tasks, including appearance editing, style transfer, scene modification, and content adaptation, enabling controlled comparison between shared source conditions and various transformation types. To validate the reliability of the benchmark, we further collect human judgment annotations and analyze their correlation with the proposed evaluation dimensions.

Using V2V-Bench, we systematically evaluate current video generation models across multiple editing scenarios and content categories. Our results show that while recent models often achieve strong perceptual quality, they still struggle to preserve source-video fidelity, maintain temporal consistency, and execute transformations with robust frame-level correspondence. These findings highlight the need for evaluation protocols that go beyond realism and semantic relevance to measure transformation correctness under source-video constraints.

Our contributions are as follows:

1) We introduce V2V-Bench, a benchmark specifically designed for video-to-video generation, which includes a hierarchical evaluation framework with 11 disentangled dimensions covering temporal alignment, structural fidelity, transformation quality, video quality, and semantic alignment.

2) We collect human judgment annotations to validate the alignment between benchmark scores and human perception.

\section{Related Work}
\label{sec:related}

\paragraph{Video generation benchmarks.}
VBench~\cite{huang2023vbench} introduced multi-dimensional evaluation for T2V generation across 16 dimensions including subject consistency, motion smoothness, and aesthetic quality. VBench-I2V extended this to image-conditioned generation with 2 additional dimensions for first-frame preservation. EvalCrafter~\cite{liu2024evalcrafter} proposed a similar multi-metric evaluation for T2V. However, all these benchmarks assume a \emph{single-input} paradigm (text or image prompt) and lack the source-output temporal alignment constraints fundamental to V2V evaluation.

\paragraph{Video editing methods.}
The V2V generation landscape spans a diverse spectrum of approaches. Early methods include instruction-based editing (InstructPix2Pix~\cite{brooks2023instructpix2pix}), attention manipulation (Prompt-to-Prompt), and one-shot tuning (Tune-A-Video~\cite{wu2023tuneavideo}). More recent advances focus on improving temporal consistency and controllability, such as feature propagation methods (TokenFlow~\cite{tokenflow2023}) and control-signal guided frameworks (ControlVideo~\cite{zhao2023controlvideo}).

Beyond single-shot editing, recent works have begun to explore \emph{frame-to-story} or narrative-driven video generation paradigms, where generation is conditioned on structured keyframes or story representations. For example, STAGE~\cite{zhang2025stage} formulates multi-shot video generation as storyboard-anchored frame pair prediction to enforce long-range consistency, while StoryAnchors~\cite{wang2025storyanchors} introduces a bidirectional framework for generating temporally coherent story frames across multiple scenes. These approaches emphasize global narrative coherence and cross-shot consistency, representing an emerging direction complementary to traditional V2V editing.

Commercial video generation models such as Grok, Runway, Gen-3, Kling, Pika, and Sora further demonstrate the practical importance of V2V capabilities~\cite{wang2025based,team2025klingavatar}. Despite these advances, existing methods make different trade-offs between edit faithfulness, temporal consistency, and source preservation, and a unified evaluation protocol for systematically comparing them remains lacking.
\section{V2V-Bench Framework}
\label{sec:method}

The proposed V2V-Bench framework consists of four sequential stages, as illustrated in Figure~\ref{fig:pipeline}.

\textbf{Stage 1: Input Preparation.}
The pipeline takes (i) a set of source videos $\{v_1, v_2, \ldots\}$ and (ii) a JSON task configuration specifying video IDs, editing prompts, and task categories. A Task Dispatcher parses the inputs and schedules editing jobs.

\textbf{Stage 2: Video Generation.}
Tasks are routed through a Model Registry that enables plug-and-play integration of heterogeneous video editing models (e.g., Veo-3.1, Grok, Open-Sora2, and user-defined models).

\textbf{Stage 3: Quality Control.}
Prior to evaluation, we first compare the generated video with its corresponding source video in terms of frame count and frame rate (FPS). If either the frame count or FPS is inconsistent, the sample is marked as a failure, indicating that the model is unable to produce outputs with the required temporal length and frame consistency. These cases are still recorded for analysis. For all remaining compliant video pairs, we proceed with evaluation across the 11 defined dimensions.

\textbf{Stage 4: Multi-Dimensional Evaluation.}
The 11 fine-grained evaluation dimensions are organized into five categories: temporal alignment, structural fidelity, transformation quality, video quality, and semantic alignment. Among these, six dimensions are specifically designed for video-to-video (V2V) evaluation: Frame Correspondence, Temporal Consistency, Structural Preservation, Layout Adherence, Edit Faithfulness, and Style Transfer Quality.

\subsection{Compliance Check}
\label{sec:constraints}

Given a source video $\mathcal{V}_s = \{s_1, \ldots, s_T\}$ with frame rate $r$ and a transformation prompt $p$, a video-to-video (V2V) model generates an output video $\mathcal{V}_o = \{o_1, \ldots, o_{T'}\}$. Unlike T2V or I2V settings, V2V requires preserving structural correspondence between input and output over time.

We define three constraints: (i) \textbf{temporal length preservation}, $T' = T$; (ii) \textbf{frame rate consistency}, $\mathrm{FPS}(\mathcal{V}_o) = r$; and (iii) \textbf{frame-level correspondence}, enforcing a mapping $o_t \leftrightarrow s_t$ for all $t \in \{1, \ldots, T\}$. The first two are enforced via a pre-evaluation compliance check, while the third is assessed through downstream metrics such as temporal alignment and structural fidelity.

\subsection{Evaluation Dimensions}
\label{sec:dimensions}

V2V-Bench evaluates 11 dimensions across 5 categories: temporal alignment, structural fidelity, transformation quality, visual quality, and semantic alignment, with 6 novel dimensions (Frame Correspondence, Temporal Consistency, Structural Preservation, Layout Adherence, Edit Faithfulness, and Style Transfer Quality) specifically designed for V2V and 4 dimensions (Motion Smoothness, Aesthetic Quality, Imaging Quality, Temporal Flickering) reused from VBench.

\subsubsection{Temporal Alignment}

This category captures the core V2V requirement: each output frame must align temporally with its source counterpart while preserving motion dynamics.

\paragraph{Frame Correspondence.}
For each frame pair $(s_t, o_t)$, we combine DINO ViT-B/16~\cite{caron2021dino} semantic features with SSIM:
\begin{equation}
S_{\mathrm{fc}}= \frac{1}{T}\sum_{t=1}^{T} \big[\alpha \cdot \cos(f_s^t, f_o^t) + (1\!-\!\alpha) \cdot \text{SSIM}(s_t, o_t)\big]
\label{eq:fc}
\end{equation}
where $f_s^t = \text{DINO}(s_t)$, $f_o^t = \text{DINO}(o_t)$, and $T = 8$ is the number of video frames used for evaluation. We set $\alpha = 0.7$ to prioritize semantic correspondence over pixel-level similarity, as V2V generation should preserve high-level content consistency rather than exact reconstruction, particularly for style transfer and appearance editing. DINO features (70\%) capture semantic/object-level alignment, while SSIM (30\%) provides complementary structural consistency. Empirically, rankings remain stable for $\alpha \in [0.65, 0.8]$, with $\alpha = 0.7$ offering a balanced trade-off between semantic robustness and structural sensitivity.

\paragraph{Temporal Consistency.}
We evaluate temporal consistency by measuring whether the generated video preserves the motion pattern of the source video.
We compare the optical flow fields of the source and generated videos using relative endpoint error:
\begin{equation}
S_{\mathrm{temp}} =
\exp \left(
-\frac{1}{T-1}
\sum_{t=1}^{T-1}
\mathbb{E}
\left[
\frac{
\|F_t^s - F_t^o\|_2
}{
\|F_t^s\|_2 + 1
}
\right]
\right),
\end{equation}
where $F_t^s$ and $F_t^o$ denote optical flow between adjacent frames in the source and generated videos.

\subsubsection{Structural Fidelity}

This category evaluates whether the output preserves the geometric structure and scene layout of the source.

\paragraph{Structural Preservation.}
Structural preservation evaluates whether object boundaries and spatial structures are retained after editing. We extract Canny edge maps from source and generated frames and compute an edge-level F1 score:
\begin{equation}
S_{\mathrm{struct}} =
\frac{1}{T}\sum_{t=1}^{T}
\mathrm{F1}\left(E(I_t^s), E(I_t^o)\right),
\end{equation}
where $E(\cdot)$ denotes edge extraction with spatial tolerance, and $I_t^s, I_t^o$ are the source and generated video frames at time $t$, respectively.

\paragraph{Layout Adherence.}
Layout adherence measures whether the global spatial arrangement of the source video is preserved. We compute frame-level structural similarity between source and generated videos:
\begin{equation}
S_{\mathrm{layout}} =
\frac{1}{T}\sum_{t=1}^{T}
\mathrm{SSIM}(I_t^s, I_t^o).
\end{equation}

\subsubsection{Transformation Quality}

This category assesses whether the V2V model correctly executed the intended transformation.

\paragraph{Edit Faithfulness.}
Edit faithfulness measures how well the generated video follows the textual editing instruction. We compute CLIP image-text similarity between sampled generated frames and the prompt:
\begin{equation}
S_{\mathrm{edit}} =
\frac{1}{T}\sum_{t=1}^{T}
\frac{\cos\left(f_{\mathrm{CLIP}}(I_t^o), f_{\mathrm{CLIP}}(p)\right)+1}{2}.
\end{equation}

\paragraph{Style Transfer Quality.}
For style-transfer tasks, we evaluate both the magnitude and direction of the style change. The magnitude is measured by VGG-19 Gram-matrix distance between source and generated frames, while the direction is measured by directional CLIP similarity:
\begin{equation}
S_{\mathrm{style}} =
\lambda \cdot M_{\mathrm{Gram}} \cdot G_{\mathrm{dir}}
+
(1-\lambda)\cdot D_{\mathrm{CLIP}},
\end{equation}
where $M_{\mathrm{Gram}}$ captures whether the style has changed and $D_{\mathrm{CLIP}}$ measures whether the change follows the target style. We assign a larger weight $\lambda = 0.6$ to Gram-based magnitude because style-transfer tasks should first exhibit a perceptible style change, while directional CLIP ensures that the change follows the requested target style. The term $G_{\mathrm{dir}}$ penalizes edits moving away from the requested direction, defined as
\begin{equation}
G_{\mathrm{dir}} =
\begin{cases}
1.0, & \text{if } \cos(\Delta I, \Delta T) \ge 0,\\
0.5, & \text{if } \cos(\Delta I, \Delta T) < 0,
\end{cases}
\end{equation}
where $\Delta I = I_{\mathrm{out}} - I_{\mathrm{src}}$ denotes the visual change direction in CLIP space, and $\Delta T = T_{\mathrm{target}} - T_{\mathrm{source}}$ denotes the desired textual style-change direction.

\subsubsection{Video Quality}

We reuse four well-validated VBench~\cite{huang2023vbench} dimensions that assess intrinsic video quality independent of the source: imaging quality, temporal flickering, aesthetic quality, and motion smoothness.
\begin{itemize}
    \item \textbf{Motion Smoothness:} Optical flow acceleration magnitude; lower acceleration indicates smoother motion trajectories.
    \item \textbf{Aesthetic Quality:} LAION aesthetic predictor~\cite{schuhmann2022laion} score reflecting color harmony, composition, and sharpness.
    \item \textbf{Imaging Quality:} BRISQUE no-reference quality assessment detecting blur, noise, and compression artifacts.
    \item \textbf{Temporal Flickering:} Mean inter-frame pixel difference; regions with natural fast motion are masked to avoid false positives.
\end{itemize}

\subsubsection{Semantic Alignment}

\paragraph{Content Preservation.}
Content preservation evaluates whether the primary visual content remains consistent with the source video. We use color-distribution similarity as a lightweight proxy:
\begin{equation}
S_{\mathrm{content}} =
\frac{1}{T}\sum_{t=1}^{T}
\mathrm{HistSim}(I_t^s, I_t^o),
\end{equation}
where $\mathrm{HistSim}$ computes the average correlation between RGB-channel histograms.

\subsection{Task Suite}
\label{sec:tasksuite}

The curated task suite contains 81 valid task instances spanning five edit video generation categories: object editing, appearance editing, style transfer, motion editing, and identity preservation. This design provides relatively even coverage of different editing objectives within a visually coherent set of human-motion videos.

In terms of video duration, the dataset mainly consists of short clips, with an average duration of approximately 8 seconds. For each video, we use Gemini-2.5-Flash to generate structured prompts. All prompts are grounded in the visual content of the source video.

\section{Experiments}
\label{sec:exp}

\subsection{Settings}

We evaluate two commercial models, Veo-3.1 and Grok-Imagine-Video, and one open-source model, Open-Sora2, on 81 editing prompts spanning style transfer, appearance editing, environment editing, identity preservation, and related tasks.

Given the broad range of possibilities with video world model testing, we also constructed a focused subset of four videos with 10 detailed task prompts. For this subset, all video generation, human judgment, and VLM-based evaluations have been completed. All generation and evaluation experiments are conducted on a single NVIDIA H100 GPU.

\subsection{Experiment 1}

\subsubsection{Compliance Check}
The compliance check is critical for video-to-video evaluation because V2V generation requires frame-level correspondence between the source and generated videos. As shown in Table~\ref{tab:compliance-check}, Veo-3.1 satisfies this requirement for all 41 evaluated samples, while Grok and Open-Sora-2 fail all cases due to frame count mismatch. This failure is primarily attributed to their inability to consistently generate videos of sufficient temporal length, which further reflects their limitations in handling long-duration video synthesis tasks. As a result, the compliance check also serves as an indicator of a model’s capability in maintaining temporal consistency over longer video generation horizons.

For Grok and Open-Sora-2, which do not pass the compliance check, we still include their outputs in the evaluation by performing frame-level alignment-based comparison, where metrics are computed over the overlapping generated frames only. This ensures a fair comparison under non-compliant generation settings while preserving evaluability across all models.

This highlights that compliance is not merely a formatting constraint, but a prerequisite for faithful V2V evaluation and for preserving the temporal structure of the input video.

\begin{table}[t]
  \centering
  \caption{Compliance check results for the three evaluated V2V models.}
  \label{tab:compliance-check}
  \small
  \setlength{\tabcolsep}{6pt}
  \begin{tabular}{lccc}
    \toprule
    \textbf{Model} & \textbf{Pass}   & \textbf{Pass Rate} & \textbf{Main Failure Mode} \\
    \midrule
    Grok        & 0 / 41   & 0.0\%   & Frame: $192 \rightarrow 185$ \\
    Veo-3.1     & 41 / 41  & 100.0\% & None \\
    Open-Sora-2 & 0 / 41   & 0.0\%   & Frame: $192 \rightarrow 129$ \\
    \bottomrule
  \end{tabular}
\end{table}

\subsubsection{Benchmark Results}

\begin{table}[t]
\centering
\caption{Three-model comparison over all 11 evaluation dimensions. Scores are averaged over 41 common video-editing tasks. Bold indicates the best and underlining indicates the second-best.}
\label{tab:v2v_all_11_dimensions}
\small
\setlength{\tabcolsep}{6pt}
\begin{tabular}{l r r r}
\toprule
\textbf{Dimension} & \textbf{Grok} & \textbf{Veo} & \textbf{Open-Sora}  \\
\midrule
Imaging Quality             & \underline{0.4979} & \textbf{0.6522} & 0.3031 \\
Temporal Flickering         & \underline{0.9836} & \textbf{0.9856} & 0.9814  \\
Structural Preservation     & \textbf{0.5726} & \underline{0.2926} & 0.2225  \\
Temporal Consistency        & \textbf{0.5289} & 0.1752 & \underline{0.3464}\\
Frame Correspondence        & \textbf{0.7895} & \underline{0.7118} & 0.6697  \\
Edit Faithfulness           & \textbf{0.6187} & \underline{0.6161} & 0.6105  \\
Aesthetic Quality           & \textbf{0.4981} & \underline{0.4976} & 0.4931 \\
Motion Smoothness           & \underline{0.9657} & \textbf{0.9865} & 0.9645  \\
Layout Adherence            & \textbf{0.7822} & 0.6564 & \underline{0.6692}\\
Style Transfer Quality      & \textbf{0.8660} & \underline{0.6903} & 0.6141  \\
Content Preservation        & \underline{0.6086} & \textbf{0.7489} & 0.4633  \\
\midrule
\textbf{Mean}               & \textbf{0.7011} & \underline{0.6376} & 0.5762  \\
\textbf{Dimensions Won}     & \textbf{7 / 11} & \underline{4 / 11} & 0 / 11 \\
\bottomrule
\end{tabular}
\end{table}

\begin{table}[t]
\centering
\caption{Comparison on the six V2V-Bench-specific dimensions.}
\label{tab:v2v_specific_6_dimensions}
\small
\setlength{\tabcolsep}{5pt}
\begin{tabular}{@{}p{0.42\columnwidth} r r r@{}}
\toprule
\textbf{V2V-Specific Dimension} & \textbf{Grok} & \textbf{Veo} & \textbf{Open-Sora} \\
\midrule
Structural Preservation & \textbf{0.5726} & \underline{0.2926} & 0.2225 \\
Temporal Consistency    & \textbf{0.5289} & 0.1752 & \underline{0.3464} \\
Frame Correspondence    & \textbf{0.7895} & \underline{0.7118} & 0.6697 \\
Edit Faithfulness       & \textbf{0.6187} & \underline{0.6161} & 0.6105 \\
Layout Adherence        & \textbf{0.7822} & 0.6564 & \underline{0.6692} \\
Style Transfer Quality  & \textbf{0.8660} & \underline{0.6903} & 0.6141 \\
\midrule
\textbf{Mean}           & \textbf{0.6937} & \underline{0.5237} & 0.5221 \\
\textbf{Dimensions Won} & \textbf{6 / 6} & 0 / 6 & 0 / 6 \\
\bottomrule
\end{tabular}
\end{table}

Table~\ref{tab:v2v_all_11_dimensions} shows that Grok achieves the highest overall mean score and wins the most dimensions, despite similar performance among models on several general metrics such as temporal flickering, aesthetic quality, and motion smoothness. In contrast, Table~\ref{tab:v2v_specific_6_dimensions} reveals larger gaps on the six V2V-specific dimensions, where Grok consistently outperforms the other models in preserving source-video structure, temporal correspondence, and edit faithfulness. This suggests that V2V-specific metrics are more discriminative than general video-quality metrics for evaluating video-to-video generation.

\subsection{Experiment 2}

We conduct a controlled study on four representative videos using 10 diverse editing tasks per video and evaluating across 11 fine-grained dimensions. Human judgments from three independent annotators assess how well the benchmark aligns with human preferences.

\begin{table}[tbp]
  \centering
  \caption{Spearman correlation between Benchmark/Human/GPT-4o/Gemini 2.5 Pro.}
  \label{tab:pooled-spearman}
  \begin{tabular}{lcc}
    \toprule
    Pair & All 11 dims  & V2V-core 6 \\
    \midrule
    Human \,$\leftrightarrow$\, \textsc{Bench}            & 0.688 & \textbf{0.905} \\
    Human \,$\leftrightarrow$\, Gemini 2.5 Pro            & 0.713 & 0.899 \\
    Human \,$\leftrightarrow$\, GPT-4o                    & \textbf{0.737} & 0.816 \\
    \midrule
    GPT-4o \,$\leftrightarrow$\, Gemini 2.5 Pro           & 0.943 & 0.912 \\
    \textsc{Bench} \,$\leftrightarrow$\, Gemini 2.5 Pro   & 0.578 & 0.826 \\
    \textsc{Bench} \,$\leftrightarrow$\, GPT-4o           & 0.578 & 0.823 \\
    \bottomrule
  \end{tabular}
\end{table}

\subsubsection{Human and VLM Preference Alignment}

To validate that the proposed evaluation method can faithfully reflect human perception, we performed human annotation for each dimension.
We show the correlation between V2V-Bench evaluation results and human preference annotations and two VLM models in Table~\ref{tab:pooled-spearman},
which shows that V2V-Bench aligns well with human judgments, achieving a Spearman correlation of $0.688$ across all 11 dimensions and a much stronger correlation of $0.905$ on the V2V-core subset. This indicates that the proposed V2V-specific dimensions better capture human preferences for source preservation, temporal correspondence, and edit fidelity than the full mixed dimensions. The two VLM judges are highly correlated with each other, but their agreement with humans is lower than that of V2V-Bench on the V2V-core dimensions.

Due to space limitations, we present the main results here and defer additional experimental results and detailed analyses to the appendix.

\section{Conclusion}
\label{sec:conclusion}
We introduce V2V-Bench, a benchmark for video-to-video generation that decomposes quality into hierarchical, disentangled dimensions, each paired with tailored prompts and dedicated evaluation methods. The benchmark spans 11 fine-grained dimensions organized into 5 categories: temporal alignment, structural fidelity, transformation quality, video quality, and semantic alignment. We curate a diverse set of source videos that cover varied scenes, motions, and visual content, each paired with editing tasks that include appearance editing, style transfer, and other challenging transformations. To validate alignment with human perception, we collect human preference annotations on prompt-based tasks across two commercial models (Grok Imagine, Gemini Veo3) and one open-source model (Open Sora 2). Benchmark results show complementary strengths: Grok Imagine achieves higher editing fidelity, while Gemini Veo3 delivers stronger visual quality. In the six dimensions specific to V2V, our benchmark achieves a Spearman correlation of $0.905$ with human judgments.

\bibliographystyle{icml2026}
\bibliography{references}

\clearpage
\appendix
\section{Additional Experimental Results}

\subsection{Benchmark Result}

\begin{table}[t]
\centering
\caption{Per-dimension scores on V2V-Bench (\textbf{higher is better}).}
\label{tab:v2v_main}
\small
\setlength{\tabcolsep}{6pt}
\begin{tabular}{l r r r}
\toprule
\textbf{Dimension} & \textbf{Veo-3.1} & \textbf{Grok} & \textbf{Open-Sora2} \\
\midrule
Imaging Quality          & 0.346             & \textbf{0.578} & \underline{0.248} \\
Temporal Flickering      & \underline{0.983} & \textbf{0.987} & 0.984           \\
Aesthetic Quality        & \textbf{0.607}    & \underline{0.508} & 0.503        \\
Motion Smoothness        & \textbf{0.983}    & \underline{0.976} & 0.970        \\
\cmidrule(lr){1-4}
Structural Preservation  & \underline{0.435} & \textbf{0.674} & 0.305           \\
Frame Correspondence     & \underline{0.711} & \textbf{0.829} & 0.638           \\
Layout Adherence         & \underline{0.565} & \textbf{0.743} & 0.548           \\
Content Preservation     & \underline{0.540} & \textbf{0.603} & 0.362           \\
\cmidrule(lr){1-4}
Temporal Consistency     & \underline{0.644} & \textbf{0.675} & 0.517           \\
\cmidrule(lr){1-4}
Edit Faithfulness        & 0.618             & \textbf{0.623} & \underline{0.618} \\
Style Transfer Quality   & 0.620             & \textbf{0.638} & \underline{0.634} \\
\midrule
\textbf{Mean (all 11)}   & \underline{0.639} & \textbf{0.705} & 0.595 \\
\textbf{Dimensions Won}  & 2 / 11 & 9 / 11 & 0 / 11 \\
\bottomrule
\end{tabular}
\end{table}

\begin{table}[t]
  \centering
  \caption{Inter-human agreement among the three annotators.}
  \label{tab:inter-human-agreement}
  \small
  \setlength{\tabcolsep}{6pt}
  \begin{tabular}{lcccc}
    \toprule
    Raters & SA & Strict $\kappa$  & DA & Decisive $\kappa$ \\
    \midrule
    A1 vs A2 & 0.554 & 0.362 & 0.783 & 0.571 \\
    A1 vs A3 & 0.546 & 0.351 & 0.786 & 0.576 \\
    A2 vs A3 & 0.960 & 0.940 & 0.982 & 0.964 \\
    \midrule
    Average  & 0.687 & 0.551 & 0.850 & 0.704 \\
    \bottomrule
  \end{tabular}
\end{table}

Table~\ref{tab:v2v_main} shows per-dimension results on four videos with carefully designed prompt-based tasks in V2V-Bench. Grok still achieves the highest mean score and wins 9 of 11 dimensions, showing strong performance in structure preservation, temporal alignment, and edit fidelity. Veo-3.1 remains competitive on general video-quality metrics, while Open-Sora2 generally lags behind.

\subsection{Inter-human Agreement}

Table~\ref{tab:inter-human-agreement} shows pairwise agreement among the three annotators. Strict agreement (SA) treats ties as a separate label, while decisive agreement (DA) considers only cases where both annotators choose a non-tie model; Cohen's $\kappa$ measures chance-corrected agreement. Decisive agreement is substantially higher, averaging $0.850$ with an average $\kappa$ of $0.704$, indicating reliable human judgments when annotators make committed preferences.

\begin{table}[t]
  \centering
  \caption{Win ratios for every dimension. V2V-core dimensions are marked with $^\dagger$.}
  \label{tab:win-ratios-full}
  \footnotesize
  \setlength{\tabcolsep}{5pt}
  \begin{tabular}{llcccc}
    \toprule
    Dimension & Model & \textsc{Bench} & Gemini 2.5 Pro & GPT-4o & Human \\
    \midrule
    \multirow{3}{*}{SP$^\dagger$}
      & Veo      & 0.450 & 0.487 & 0.550 & 0.388 \\
      & Grok     & \textbf{0.963} & \textbf{0.650} & \textbf{0.588} & \textbf{0.896} \\
      & OpenSora & 0.087 & 0.362 & 0.362 & 0.217 \\
    \midrule
    \multirow{3}{*}{TC$^\dagger$}
      & Veo      & 0.662 & 0.463 & 0.525 & 0.446 \\
      & Grok     & \textbf{0.800} & \textbf{0.650} & \textbf{0.600} & \textbf{0.840} \\
      & OpenSora & 0.037 & 0.388 & 0.375 & 0.215 \\
    \midrule
    \multirow{3}{*}{FC$^\dagger$}
      & Veo      & 0.425 & 0.487 & 0.537 & 0.404 \\
      & Grok     & \textbf{0.900} & \textbf{0.637} & \textbf{0.600} & \textbf{0.898} \\
      & OpenSora & 0.175 & 0.375 & 0.362 & 0.198 \\
    \midrule
    \multirow{3}{*}{EF$^\dagger$}
      & Veo      & 0.388 & 0.537 & 0.537 & 0.528 \\
      & Grok     & \textbf{0.650} & \textbf{0.594} & \textbf{0.575} & \textbf{0.734} \\
      & OpenSora & 0.463 & 0.369 & 0.388 & 0.237 \\
    \midrule
    \multirow{3}{*}{LA$^\dagger$}
      & Veo      & 0.338 & 0.500 & \textbf{0.562} & 0.381 \\
      & Grok     & \textbf{0.850} & \textbf{0.637} & 0.550 & \textbf{0.923} \\
      & OpenSora & 0.312 & 0.362 & 0.388 & 0.196 \\
    \midrule
    \multirow{3}{*}{CP$^\dagger$}
      & Veo      & 0.512 & 0.525 & 0.550 & 0.392 \\
      & Grok     & \textbf{0.713} & \textbf{0.625} & \textbf{0.588} & \textbf{0.879} \\
      & OpenSora & 0.275 & 0.350 & 0.362 & 0.229 \\
    \midrule
    \multirow{3}{*}{AQ}
      & Veo      & \textbf{1.000} & 0.475 & 0.537 & \textbf{0.789} \\
      & Grok     & 0.463 & \textbf{0.625} & \textbf{0.575} & 0.475 \\
      & OpenSora & 0.037 & 0.400 & 0.388 & 0.236 \\
    \midrule
    \multirow{3}{*}{MS}
      & Veo      & \textbf{0.706} & 0.450 & 0.525 & \textbf{0.665} \\
      & Grok     & 0.381 & \textbf{0.662} & \textbf{0.600} & 0.640 \\
      & OpenSora & 0.412 & 0.388 & 0.375 & 0.196 \\
    \midrule
    \multirow{3}{*}{STQ}
      & Veo      & 0.000 & 0.475 & 0.537 & \textbf{0.527} \\
      & Grok     & 0.625 & \textbf{0.637} & \textbf{0.575} & 0.502 \\
      & OpenSora & \textbf{0.875} & 0.388 & 0.388 & 0.471 \\
    \midrule
    \multirow{3}{*}{IM}
      & Veo      & 0.412 & 0.438 & 0.537 & \textbf{0.843} \\
      & Grok     & \textbf{0.850} & \textbf{0.650} & \textbf{0.575} & 0.406 \\
      & OpenSora & 0.237 & 0.412 & 0.388 & 0.250 \\
    \midrule
    \multirow{3}{*}{TF}
      & Veo      & 0.338 & 0.388 & 0.525 & 0.415 \\
      & Grok     & \textbf{0.675} & \textbf{0.675} & \textbf{0.588} & \textbf{0.588} \\
      & OpenSora & 0.487 & 0.438 & 0.388 & 0.498 \\
    \bottomrule
  \end{tabular}
\end{table}

\subsection{Win Ratio}

Given the human preference and VLM annotations, we compute the win ratio for each model on all the dimensions. In each pairwise comparison, a model receives a score of 1 if its generated video is preferred by annotators, while the other model receives a score of 0. In the case of a tie, both models receive a score of 0.5. The win ratio of each model is then defined as the total score accumulated across all pairwise comparisons divided by the total number of comparisons in which the model participates.

\begin{table*}[!t]
\centering
\caption{Prompts for the four task types.}
\label{tab:prompt_tasks}
\small
\setlength{\tabcolsep}{5pt}
\renewcommand{\arraystretch}{1.15}
\begin{tabular}{c l p{0.72\textwidth}}
\toprule
\textbf{\#} & \textbf{Task Type} & \textbf{Prompt} \\
\midrule
2 & Replace Object &
Replace the large yellow surfboard held by the man with an identical-shaped deep ocean blue surfboard. \\
4 & Change Lighting &
Transform the office lighting from bright neutral daylight to warm golden-hour evening lighting. \\
7 & Change Background &
Replace the dark, dimly lit office background with a bright modern open-plan office environment. \\
5 & VFX (Visual Effects) &
Apply a vivid bioluminescent ocean glow effect to the wave in the surf video. \\
\bottomrule
\end{tabular}
\end{table*}

\begin{figure*}[!t]
\centering
\setlength{\abovecaptionskip}{4pt}
\setlength{\belowcaptionskip}{0pt}

\begin{subfigure}{0.235\textwidth}
    \centering
    \includegraphics[width=\linewidth]{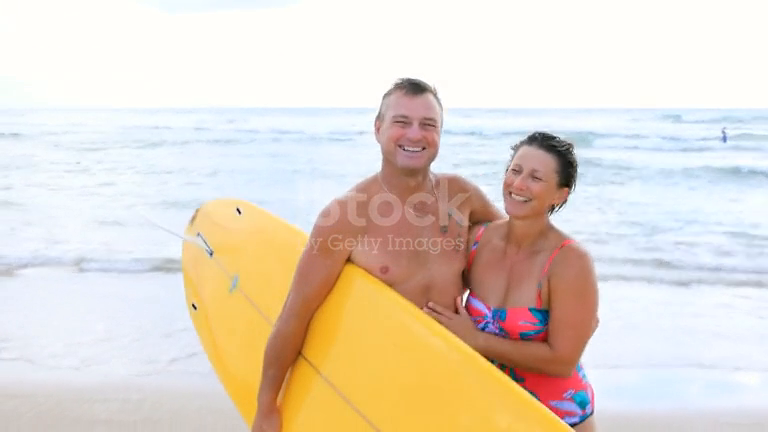}
\end{subfigure}\hfill
\begin{subfigure}{0.235\textwidth}
    \centering
    \includegraphics[width=\linewidth]{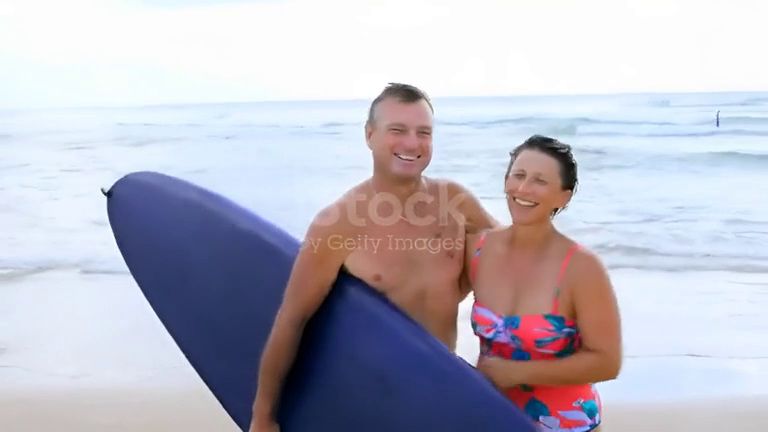}
\end{subfigure}\hfill
\begin{subfigure}{0.235\textwidth}
    \centering
    \includegraphics[width=\linewidth]{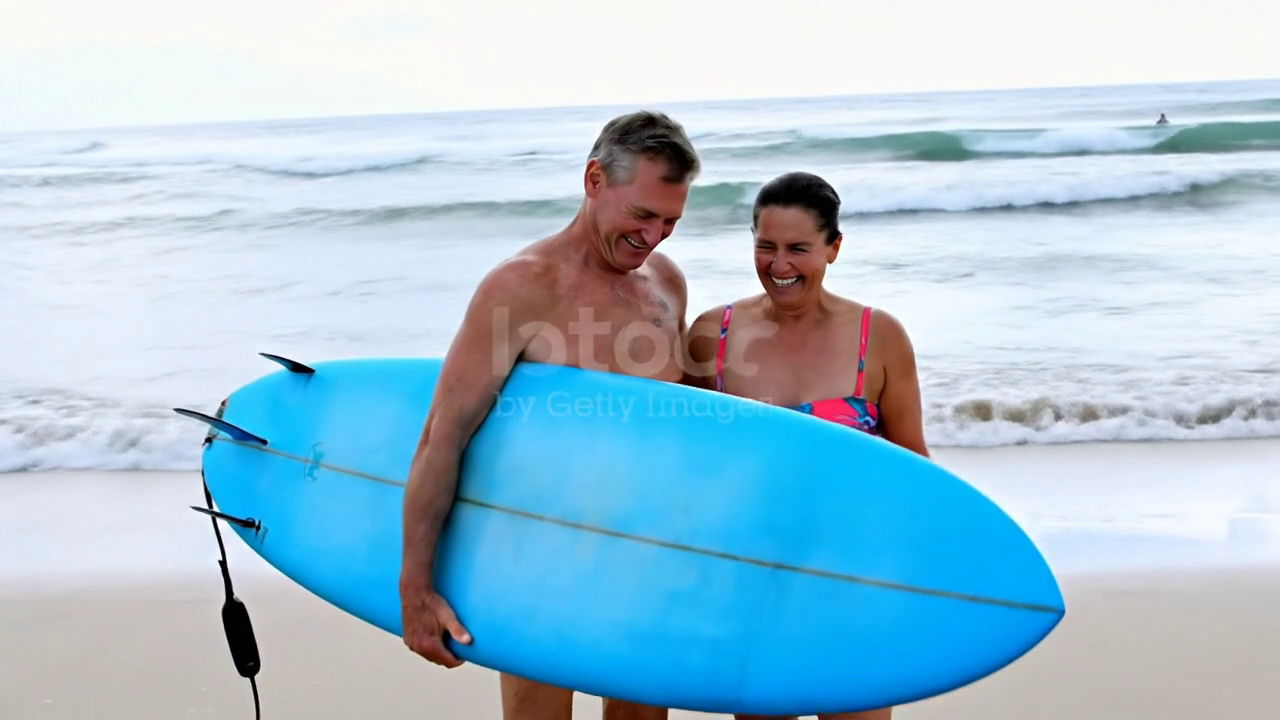}
\end{subfigure}\hfill
\begin{subfigure}{0.235\textwidth}
    \centering
    \includegraphics[width=\linewidth]{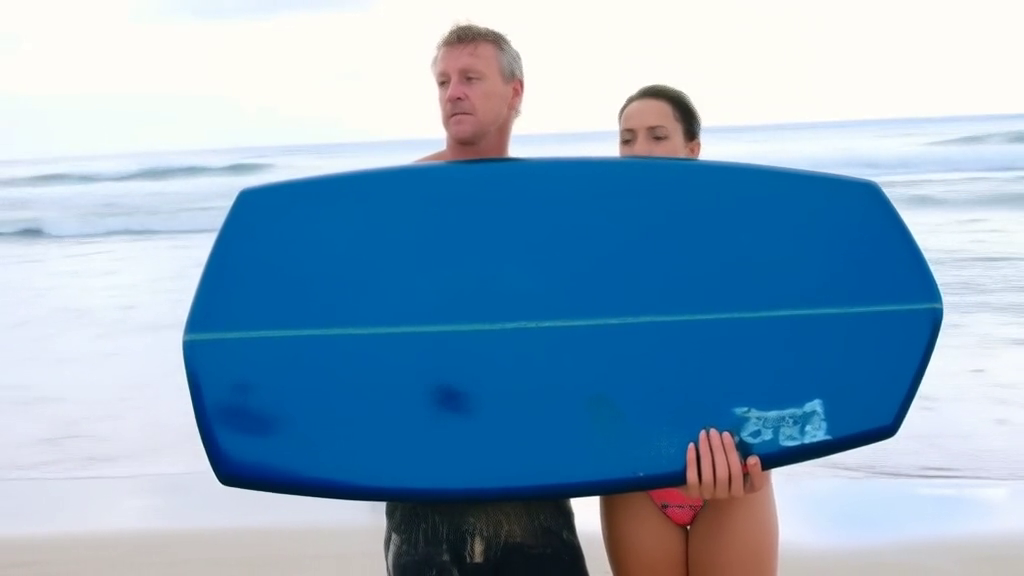}
\end{subfigure}

\vspace{0.35em}

\begin{subfigure}{0.235\textwidth}
    \centering
    \includegraphics[width=\linewidth]{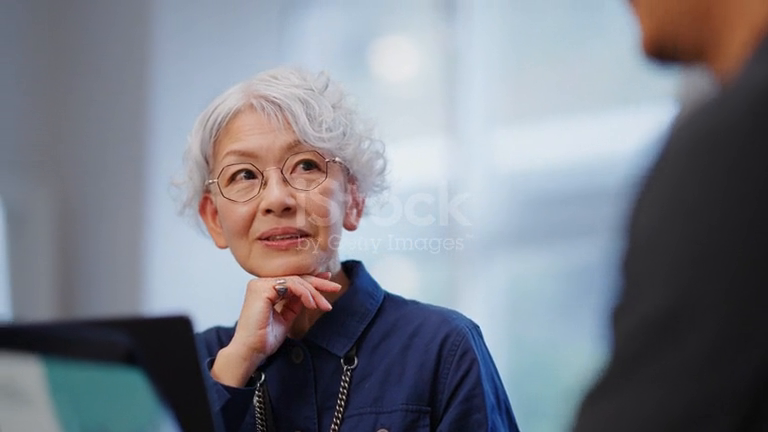}
\end{subfigure}\hfill
\begin{subfigure}{0.235\textwidth}
    \centering
    \includegraphics[width=\linewidth]{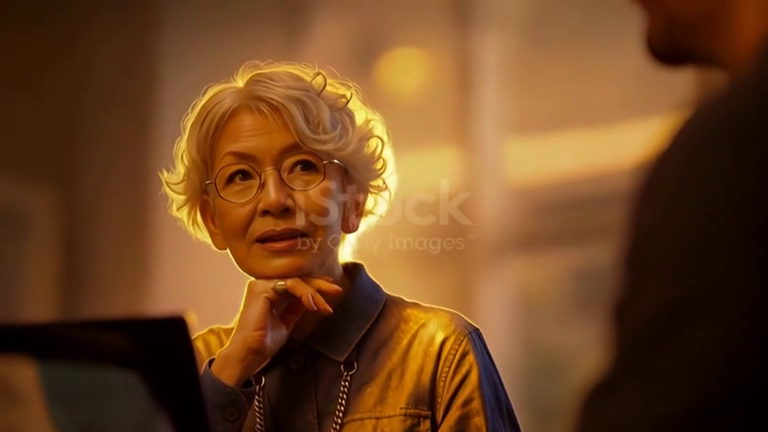}
\end{subfigure}\hfill
\begin{subfigure}{0.235\textwidth}
    \centering
    \includegraphics[width=\linewidth]{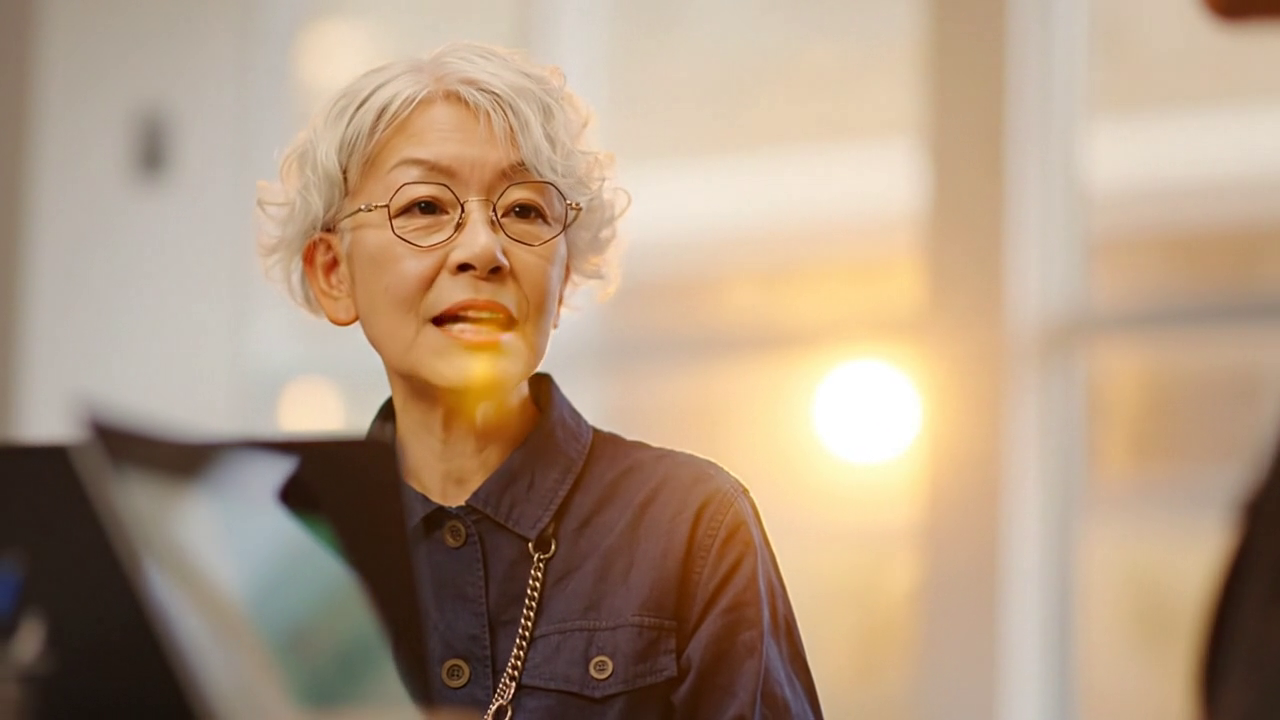}
\end{subfigure}\hfill
\begin{subfigure}{0.235\textwidth}
    \centering
    \includegraphics[width=\linewidth]{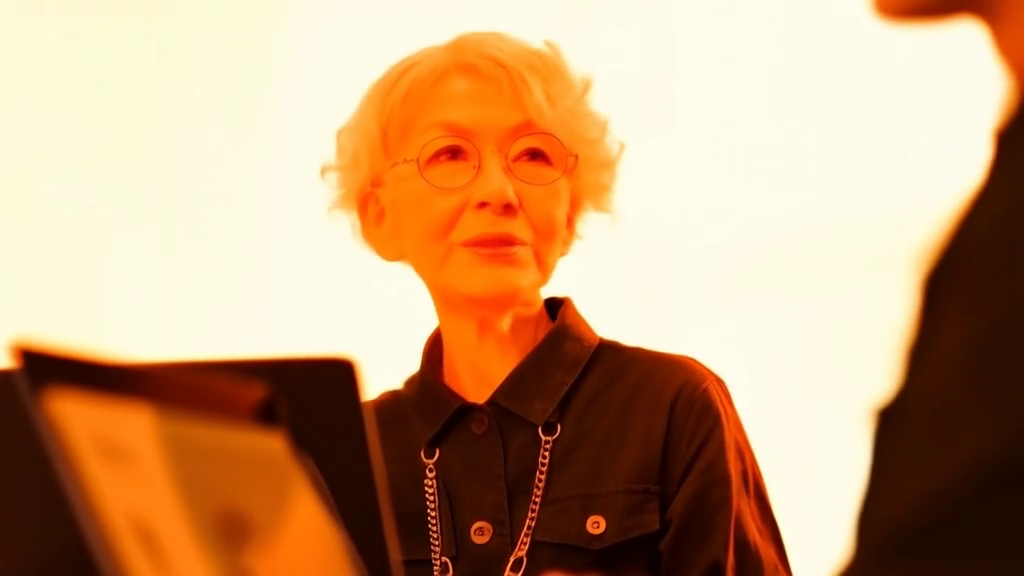}
\end{subfigure}

\vspace{0.35em}

\begin{subfigure}{0.235\textwidth}
    \centering
    \includegraphics[width=\linewidth]{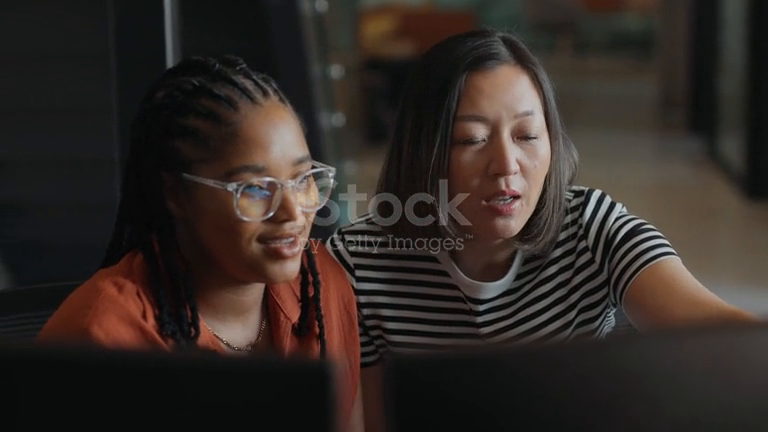}
\end{subfigure}\hfill
\begin{subfigure}{0.235\textwidth}
    \centering
    \includegraphics[width=\linewidth]{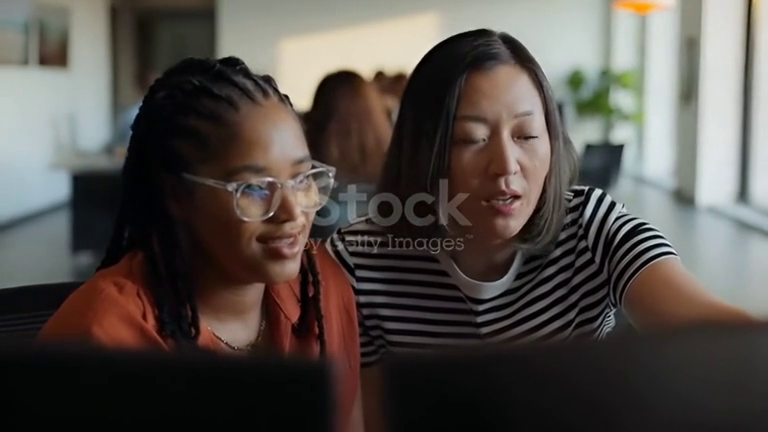}
\end{subfigure}\hfill
\begin{subfigure}{0.235\textwidth}
    \centering
    \includegraphics[width=\linewidth]{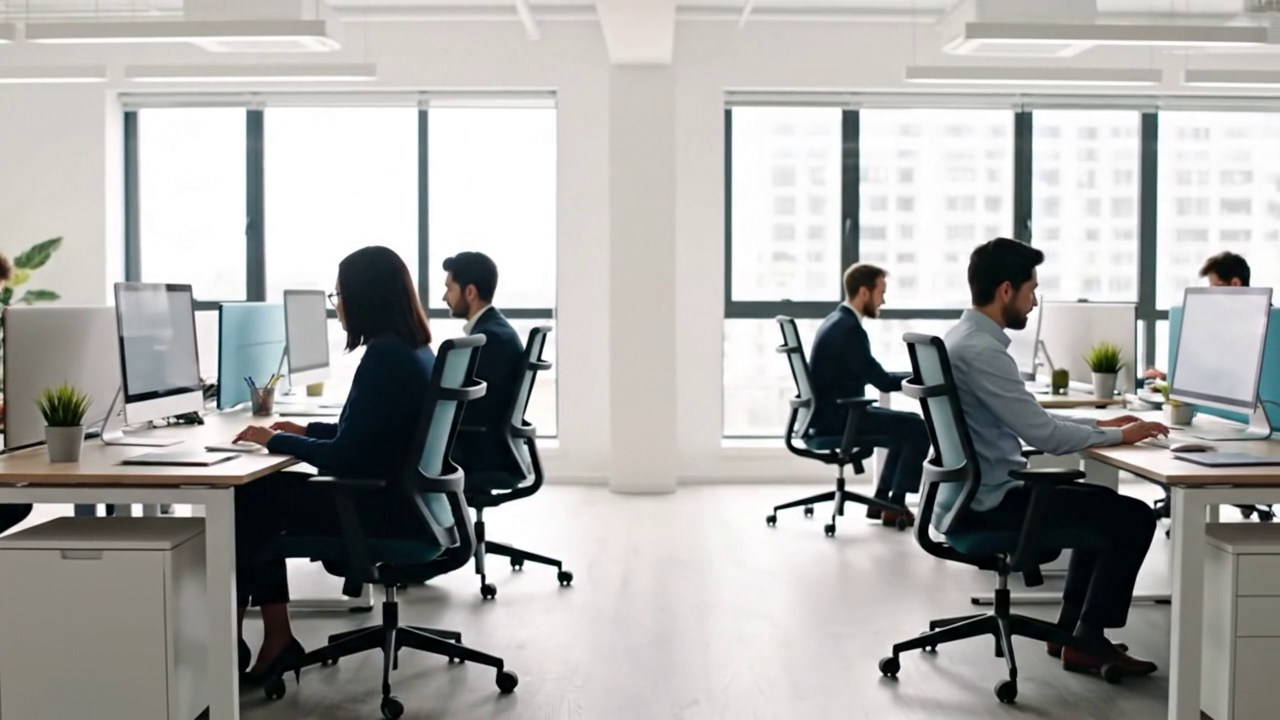}
\end{subfigure}\hfill
\begin{subfigure}{0.235\textwidth}
    \centering
    \includegraphics[width=\linewidth]{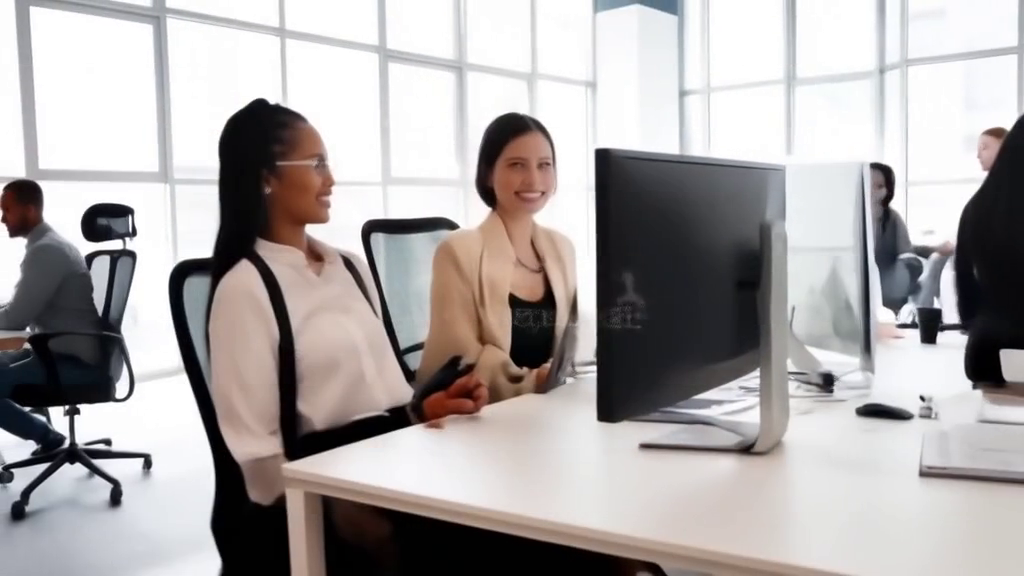}
\end{subfigure}

\vspace{0.35em}

\begin{subfigure}{0.235\textwidth}
    \centering
    \includegraphics[width=\linewidth]{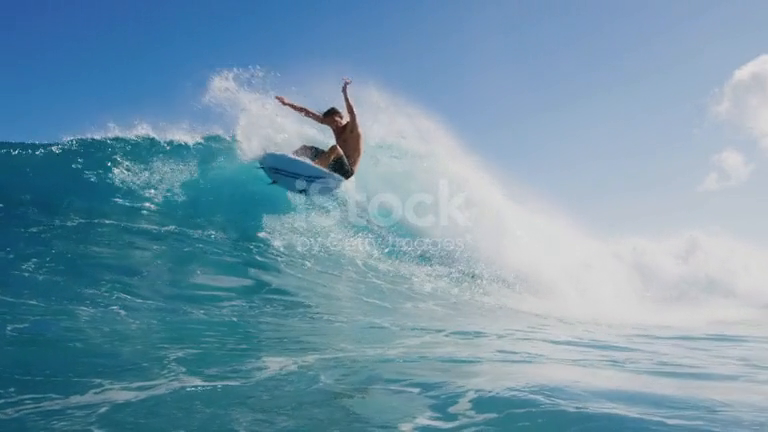}
\end{subfigure}\hfill
\begin{subfigure}{0.235\textwidth}
    \centering
    \includegraphics[width=\linewidth]{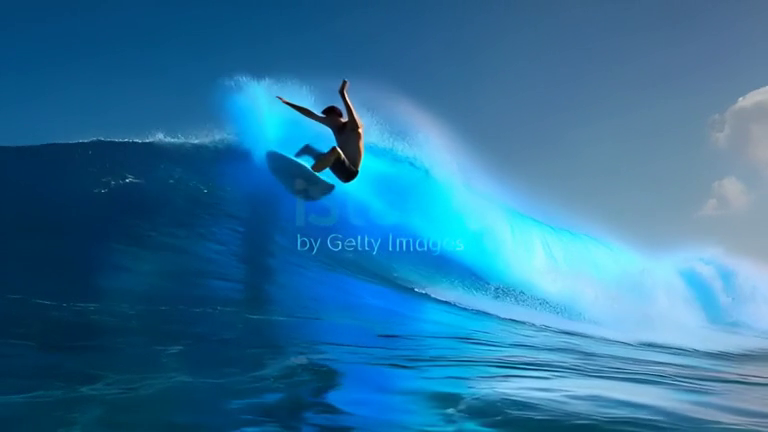}
\end{subfigure}\hfill
\begin{subfigure}{0.235\textwidth}
    \centering
    \includegraphics[width=\linewidth]{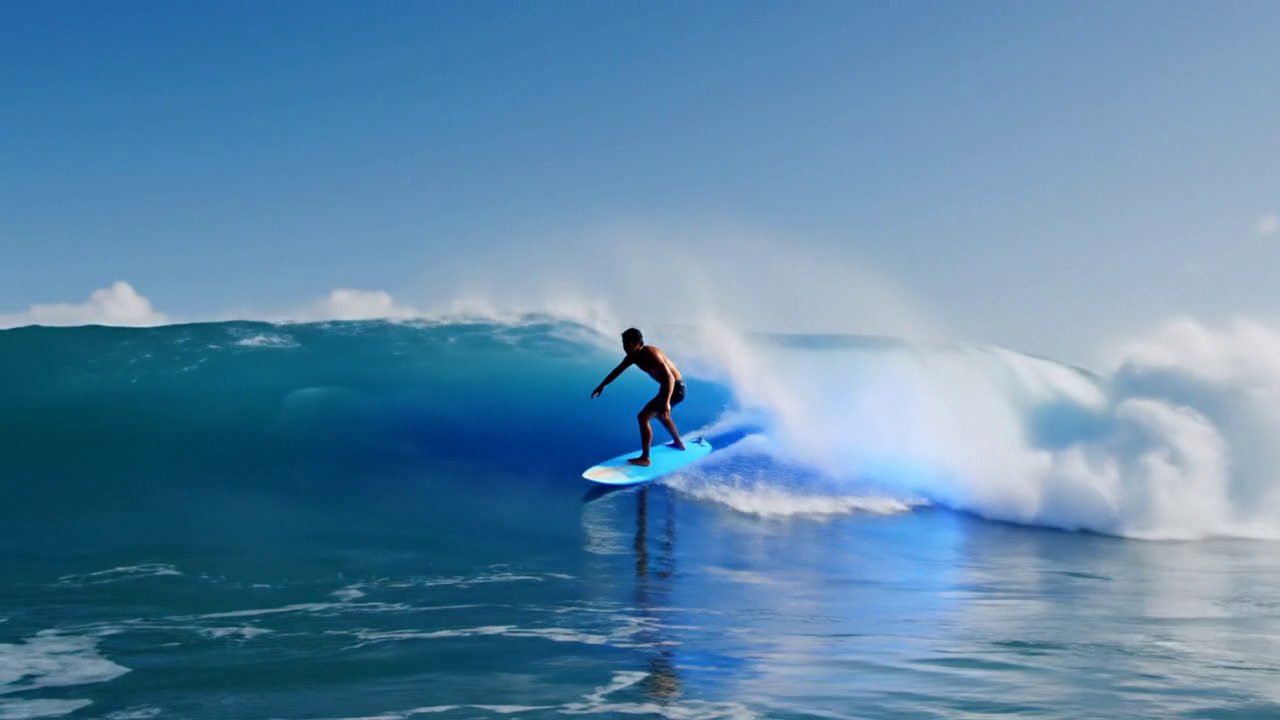}
\end{subfigure}\hfill
\begin{subfigure}{0.235\textwidth}
    \centering
    \includegraphics[width=\linewidth]{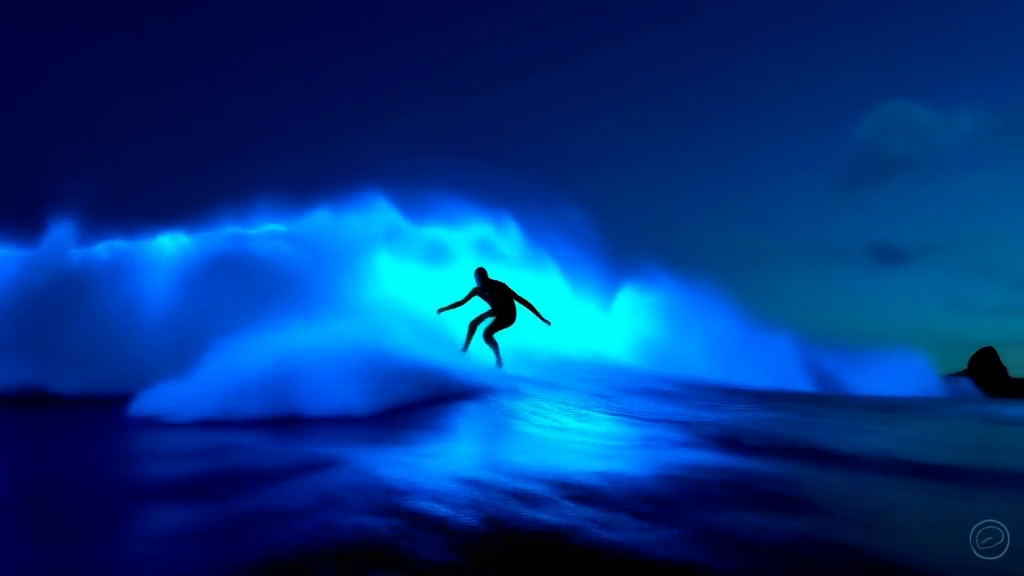}
\end{subfigure}

\caption{Qualitative comparison across different tasks. Columns from left to right show Raw, Grok, Veo-3.1, and Open-Sora2. Rows from top to bottom correspond to Tasks 2, 4, 7, and 5.}
\label{fig:8images}
\end{figure*}

Table~\ref{tab:win-ratios-full} reports the pairwise win ratios for each model across evaluation dimensions and annotation sources. On the V2V-core dimensions, both \textsc{Bench} and human judgments consistently favor Grok, especially for structural preservation, frame correspondence, layout adherence, and content preservation, indicating stronger source-video fidelity and temporal alignment. In contrast, the general quality dimensions show more source-dependent variation: Veo performs strongly in human judgments for aesthetic quality, motion smoothness, and imaging quality, while Grok remains preferred by the automatic benchmark on several perceptual metrics. The VLM judges generally produce more compressed win-ratio distributions than human judgments and \textsc{Bench}, suggesting that they are less discriminative in separating model quality. Overall, the table shows that V2V-Bench closely reflects human preferences on the V2V-specific dimensions where source preservation and edit fidelity are most critical.

\subsection{Visual Result}

Figure~\ref{fig:8images} shows qualitative comparisons across tasks. From left to right, we present the 120th frame of the source video and the outputs from Grok, Veo-3.1, and Open-Sora-2. Rows correspond to Replace Object, Change Lighting, Change Background, and VFX. The prompts for each task are in Table~\ref{tab:prompt_tasks}.

Grok consistently preserves structural fidelity, maintaining spatial relationships and human pose while applying edits. In contrast, Veo-3.1 and Open-Sora-2 exhibit structural drift in background and VFX tasks, often deviating significantly from the source. For object replacement, Veo-3.1 modifies the object but alters pose, while Open-Sora-2 introduces larger deviations in both pose and expression.

\end{document}